\documentclass[10pt, a4paper]{article}
\usepackage{lrec}
\usepackage{graphicx}
\usepackage{tabularx}
\usepackage{soul}

\usepackage{xcolor}

\usepackage{todonotes}
\usepackage{epstopdf}
\usepackage[latin1]{inputenc}

\usepackage{hyperref}
\usepackage{xstring}

\usepackage{multirow}
\usepackage{multicol}
\usepackage{subcaption}

\usepackage{verbatim}
\newcommand{\secref}[1]{\StrSubstitute{\getrefnumber{#1}}{.}{ }}

\title{Detecting Multiword Expression Type Helps Lexical Complexity Assessment}

\name{Ekaterina Kochmar,$^1$ Sian Gooding,$^1$ Matthew Shardlow$^2$}

\address{$^1$ University of Cambridge, Department of Computer Science and Technology \\
         $^2$ Manchester Metropolitan University\\
         $\{$ek358, shg36$\}$@cam.ac.uk, m.shardlow@mmu.ac.uk\\}

\abstract{Multiword expressions (MWEs) represent lexemes that should be treated as single lexical units due to their idiosyncratic nature. Multiple NLP applications have been shown to benefit from MWE identification, however the research on lexical complexity of MWEs is still an under-explored area. In this work, we re-annotate the Complex Word Identification Shared Task 2018 dataset of~\newcite{yimam-etal-2017-cwig3g2}, which provides complexity scores for a range of lexemes, with the types of MWEs. We release the MWE-annotated dataset with this paper, and we believe this dataset represents a valuable resource for the text simplification community. In addition, we investigate which types of expressions are most problematic for native and non-native readers. Finally, we show that a lexical complexity assessment system benefits from the information about MWE types. \\ \newline  }

\begin{document}

\maketitleabstract

\section{Introduction}



Complex word identification (CWI) is a well-established task, with applications in text complexity assessment and lexical simplification \cite{paetzold-2015-reliable,saggion2017automatic}. CWI is concerned with the identification of words in need of simplification and is often considered the first step in a lexical simplification pipeline~\cite{shardlow-2013-comparison}. For example, a CWI system may identify {\em appreciate} as complex in:

\vspace{-0.3em}
\begin{quote}
It made me \textit{appreciate} freedom
\end{quote}
\vspace{-0.3em}

A lexical simplification system may then suggest replacing {\em appreciate} with {\em value}, making the new sentence easier to understand for the intended reader. Most research to date has focused on complexity at the level of individual words only, despite the fact that complexity often relates to whole chunks of text. Take the following sentence for example:
\vspace{-0.3em}
\begin{quote}
Protesters used \textit{sledge hammers} to tear apart the security wall
\end{quote}
\vspace{-0.3em}
In a traditional lexical simplification pipeline, a CWI component may identify the word {\em sledge} as complex, and a lexical simplifier may then try to replace {\em sledge}, for example, with {\em sleigh}. However in this sentence {\em sledge} occurs as part of an expression {\em sledge hammers}, therefore a system tasked with {\em lexical complexity assessment} should instead identify {\em sledge hammers} as a single lexical unit, assess its complexity as such and, if necessary, attempt to simplify it as a whole (for instance, to {\em lump hammers}). 

{\em Sledge hammers} is an example of a multiword expression (MWE) -- an expression which is made up of at least two words and which has idiosyncratic interpretation that crosses word boundaries or spaces~\cite{sag-et-al-2002}. Due to this distinctive nature, many areas in NLP, including parsing~\cite{constant-2017}, machine translation~\cite{constant-2017,carpuat-diab-2010}, keyphrase/index term extraction~\cite{newman-etal-2012}, and language acquisition research~\cite{ellis-etal-2008}, benefit from treating MWEs as single lexical units.

In this paper, we argue that lexical complexity assessment systems should also treat MWEs as single units and assess their complexity as a whole, rather than on a word-by-word basis. In addition, identifying the type of the MWE is key to knowing how to simplify it. Consider the following sentence as an example:
\vspace{-0.3em}
\begin{quote}
    Thousands of protesters faced off against \textit{Interior Ministry} troops
\end{quote}
\vspace{-0.3em}
A lexical complexity assessment system might identify that {\em Interior Ministry} is an MWE in need of simplification, and that simplification would need to include the whole phrase. Knowing that {\em Interior Ministry} is a multiword named entity, the simplification system may also recognize that the most successful strategy at simplifying this expression would require providing an explanation or pointing a reader at a Wikipedia page, rather than searching for an appropriate synonym.

To date, two shared tasks on CWI have been organized~\cite{paetzold-specia-2016-semeval,yimam-etal-2018-report}, with participating systems typically focusing on identifying complexity through supervised learning. The 2018 shared task on CWI~\cite{yimam-etal-2018-report} used a dataset by~\newcite{yimam-etal-2017-cwig3g2} of $34879$ simple and complex lexemes with annotations encoding binary complex/simple decisions as well as representing the proportion of $20$ annotators that found the lexeme to be complex. These lexemes covered both single tokens ($30147$) and ``phrases" ($4732$) --- sequences longer than one word selected by the annotators. The proportion of ``phrases" in this dataset amounts to $\approx$$13$$-$$14\%$ depending on the particular data split, however none of the participating teams addressed complex phrase detection specifically. The top performing system at the competition~\cite{gooding-kochmar-2018-camb} noted that during training they were able to get the best performance by simply assigning any ``phrase'' to the complex class, rather than assessing its complexity in a focused way.

In this work, we address the task of complexity assessment for MWEs, and re-annotate the ``phrases" from the CWI Shared Task 2018 with respect to their MWE status and type (Section~\ref{sec:annotation}). This allows us to draw conclusions about the complexity of each MWE type for native and non-native readers as well as compare complexity of different types. We show that there is great variation in the complexity of MWE types and we also demonstrate that incorporating the type of MWE into a lexical complexity assessment system improves its performance (Section~\ref{sec:baseline}).

In this work, we make the following contributions:
\begin{enumerate}\vspace{-0.3em}
    \item We annotate and release a dataset of multiword expressions based on the CWI Shared Task 2018 dataset~\cite{yimam-etal-2017-cwig3g2}.\footnote{The dataset is available at \url{https://github.com/ekochmar/MWE-CWI} under CC-BY 4.0 license.} Together with the original complexity labels, this dataset represents a valuable resource for the text simplification community.\vspace{-0.3em}
    \item We explore and report statistics on which types of expressions are most problematic for native and non-native readers.\vspace{-0.3em}
    \item Finally, we show that a lexical complexity assessment system benefits from the information about the presence and type of an MWE.
\end{enumerate}

\section{Background}


\subsection{Complexity Assessment and Simplification}
Complex word identification has traditionally been approached through one of three types of methods: {\em simplify-all} aimed at simplifying every token and keeping only the changes resulting in actual simplification; {\em threshold-based} methods applying pre-defined thresholds to one or more measures (e.g., lexical frequency, word length, etc.); and {\em supervised learning-based} methods~\cite{shardlow-2013-comparison}. Recent approaches in supervised learning have covered sequence labelling for complex word identification~\cite{gooding-kochmar-2019-complex}, the use of neural networks such as CNNs~\cite{aroyehun-etal-2018-complex}, and work on feature-based approaches such as character n-grams~\cite{popovic-2018-complex}.

To date, two shared tasks on complex word identification have been organised: the shared task in 2016 was co-located with SemEval~\cite{paetzold-specia-2016-semeval}, and the shared task in 2018 was co-located with the Workshop on Innovative Use of NLP for Building Educational Applications~\cite{yimam-etal-2018-report}. These workshops have served to drive recent research in CWI, providing new datasets for the community and giving insights on what techniques work well. In both tasks, supervised feature-based approaches to CWI scored highly~\cite{paetzold-specia-2016-sv000gg,gooding-kochmar-2018-camb}. In our work, we use the English portion of the dataset from the CWI Shared Task 2018~\cite{yimam-etal-2017-cwig3g2}.

Despite the fact that the shared tasks attracted attention to complexity assessment and provided the research community with valuable data, the research on lexical complexity of MWEs is still an under-explored area. Most previous work has focused on assessing the complexity of single words, with a few notable exceptions: for instance, work on metaphor identification in simplification~\cite{clausen-nastase-2019-metaphors} and work on creating tables of paraphrases~\cite{maddela-xu-2018-word} that can be used to simplify medical terminology~\cite{shardlow-nawaz-2019-neural}. Our work fills a gap that is left in understanding and identifying the complexity of MWEs.

\subsection{Multiword Expressions}

Multiword expressions are longer than one word and show idiosyncratic behaviour in terms of syntax and/or semantics. MWEs are pervasive in language: for instance,~\newcite{jackendoff1997} estimates that the number of MWEs in a speaker's lexicon is comparable to that of single words. Identification of the broad variety of MWEs in language is, however, a non-trivial task~\cite{sag-et-al-2002}.

 Linguists distinguish between {\em lexicalized phrases} like {\em kick the bucket}, and {\em institutionalized phrases} like {\em traffic lights}~\cite{bauer1983,sag-et-al-2002}. The former are characterized by at least partially idiosyncratic syntax or semantics, the latter are syntactically and semantically compositional, but are common phrases.

Typically, the distinction between the two groups of MWEs is drawn on the basis of compositionality  and ``substitutionability": despite {\em traffic lights} being semantically transparent, its components cannot be freely substituted with synonyms without distortion of the original meaning or violation of language conventions. In general, compositionality and the strength of association between words in MWEs range from fully transparent collocations to completely opaque idioms~\cite{hermann-etal-2012}, which adds to the complexity of the task of MWE identification. At the {\em collocation} end of this spectrum lie expressions consisting of statistically significant co-occurrences of words, which are predictably frequent because of real world events or other non-linguistic factors~\cite{sag-et-al-2002}. Unlike lexicalized and institutionalized phrases, individual words within collocations can be replaced with their synonyms without violating the meaning or language conventions: for instance, {\em steep fall} can be partially substituted with {\em sharp decline} or variations thereof.

Previous research introduced the notion of {\em strong} MWEs for lexicalized and institutionalized phrases and {\em weak} MWEs for more transparent and flexible collocations~\cite{schneider-et-al-2014}. We focus on the annotation of various types of lexicalized and institutionalized phrases, leaving collocations out since they can be simplified on a word-by-word basis. As the primary goal of our research is to identify MWEs that might be deemed complex by readers and will need to be simplified as a single unit, when annotating MWEs in data we pose two questions:
\vspace{-0.5em}
\begin{enumerate}
    \item Might an expression be deemed complex as a whole?\vspace{-0.5em}
    \item Should it be simplified as a single lexical unit rather than on a word-by-word basis?
\end{enumerate}

\section{Data and Annotation} \label{sec:annotation}


The CWI Shared Task 2018 dataset~\cite{yimam-etal-2017-cwig3g2} is the most comprehensive dataset of complex words and ``phrases" annotated in context. The dataset covers three text genres (professionally written {\sc News}, {\sc WikiNews} written by amateurs, and {\sc Wikipedia} articles) annotated by $10$ native and $10$ non-native English speakers via Amazon Mechanical Turk. Annotators were presented with text passages ($5$$-$$10$ sentences) and asked to select up to $10$ words or sequences of words that they deemed complex. There were no restrictions on the types of words or sequences that the annotators could select except that annotators were not allowed to select function words like determiners and numbers, and phrases of more than $50$ characters in length. 

Each paragraph was annotated by all annotators and presented in two formats: under the {\em binary} setting, a lexeme received $1$ if any annotator selected it as complex, under the {\em probabilistic} setting, the proportion of annotators who marked a lexeme as complex was used as a label on a scale of $[0.0, 1.0]$ with a step of $0.05$. For example, a complexity value of $0.15$ for {\em Interior Ministry} indicates that $3$ out of $20$ annotators selected this ``phrase" as complex in context.


In the original CWI annotation scheme, lexemes with a complexity value of $0$ represent both content words and ``phrases" that were not selected as complex by any annotators. Although the procedure for simple word extraction is straightforward, as one may simply include all content words not explicitly selected by the annotators, the procedure for simple ``phrase'' extraction is less clear as the variation of ``phrases'' that one can automatically extract from data is prohibitively large. Data inspection shows that the simple ``phrases'' in the dataset represent text chunks rather than MWEs selected in a focused way.\footnote{Examples include fully productive compositional expressions like {\em his drive}, sentence fragments like {\em then heard}, etc.}  
 As about $79\%$ of ``phrases" are annotated as complex, with the vast majority ($43\%$) annotated as complex by a single annotator, a simple strategy of outputting $1$ as the binary prediction and $0.05$ as the probabilistic score proves to yield better results than predicting ``phrase" complexity score in any more sophisticated way~\cite{gooding-kochmar-2018-camb}.

The CWI Shared Task 2018 dataset represents a valuable resource for research on lexical complexity assessment and lexical simplification, but since the annotators of the original dataset were not tasked with annotating MWEs and were allowed to select any sequence of words up to $50$ characters in length, we argue that this dataset benefits from further MWE-focused annotation. Therefore, we first set out to re-annotate all ``phrases" from the CWI Shared Task 2018 dataset. In particular, we focus on (a) annotating whether a ``phrase" from the original dataset is an MWE or not, and if it is (b) which type of an MWE it represents. We have \textbf{not} re-annotated this data for complexity --- instead we reuse the original (binary and probabilistic) complexity labels from the shared task.

\subsection{Annotation Scheme}

We adopted the MWE categorization framework formulated by~\newcite{schneider-et-al-2014}. This framework covers a wide variety of MWE types including both lexicalized (most types in the scheme) and institutionalized (subset of multiword compounds) expressions. The annotations were performed by the three authors of this paper, all trained in linguistics and NLP. We ran the annotation in a series of rounds, where the original scheme of~\newcite{schneider-et-al-2014} was used in its unadopted form for the first round of annotating $100$ examples from the dataset only. As a result of resolving disagreements and discussing the task after the first round of annotation, a set of guidelines was developed and followed in subsequent rounds of annotation. Inter-annotator agreement was assessed after each round to ensure consistency. 

We made the following modifications to the original scheme:

\begin{description}
\item[Not MWE:] As the dataset we annotate in this work contains sequences of words selected by the annotators which do not always constitute an MWE, we use category {\tt not MWE} for such cases. Examples include {\em authorities should annul the}, {\em IP address is blocked}, etc.
\item[Not MWE but contains MWE(s)] is reserved for the sequences of words that do not constitute an MWE in full but contain MWE(s) as part of the expression: examples include combinations of several MWEs as in {\em \underline{Clarinet Concerto} and \underline{Clarinet Quintet}}, combinations of qualifiers and MW compounds as in {\em collapsed \underline{property sector}}, and similar cases.
\item[Merge of verb-particle and other phrasal verb categories:] We reason that, from a simplification point of view, the two original categories are not distinct enough and from the linguistic point of view it is hard to make clear distinction during annotation. Examples include {\em close down}, {\em go about} and similar constructions.
\item [Deprecation of phatic and proverb categories:] We found no examples of these categories in our data, and we do not report on these in our analysis. Our data is based on Wikipedia, News and WikiNews articles which are unlikely to contain these more informal expression types.
\end{description}

Table \ref{tab:categories} presents the full list of categories used in our annotation with descriptions of the types, examples and suggested directions for simplification. For brevity, we use the term {\em conventionalized} to denote semantically, syntactically or statistically idiosyncratic expressions, i.e. whenever the type may cover both lexicalized and institutionalized MWEs. Throughout the annotation process, we maintained a set of annotation guidelines, which we updated regularly with clarifications as we met to discuss our annotations. The guidelines are included in the data release.

\newcommand{\specialcell}[2][l]{%
  \begin{tabular}[#1]{@{}l@{}}#2\end{tabular}}
  
\begin{table*}
\begin{center}
\begin{tabular}{|m{2.8cm}|p{4.55cm}|l|p{4.55cm}|}
      \hline
      \multicolumn{1}{|c|}{\textbf{MWE Type}} 
      & \multicolumn{1}{c|}{\textbf{Description}} 
      & \multicolumn{1}{c|}{\textbf{Examples} }
      & \multicolumn{1}{c|}{\textbf{Proposed Simplification}} \\
      \hline
        MW named entities 
        & Concrete and unique named entities, which refer to people, organizations, etc.  
        & \specialcell{{\em Alawite sect} \\ {\em Formica Fusca} }
        & Link to a description, ontology or encyclopedia page  
        \\\hline
        
        MW compounds 
        & Conventionalized expressions with a clear meaning extending that of the individual tokens; include compound nominals. Often have a dictionary entry.
        & \specialcell{ {\em life threatening} \\{\em property sector}}
        & Replace full MWE with a simpler word or MWE.
        \\ \hline
        
        Verb-particle and other phrasal verbs 
        & Multiword verbal expressions, consisting of a verb typically attaching a particle or an adverb.
        & \specialcell{{\em close down} \\{\em get rid of}}
        & Replace full MWE with a simpler verb or MWE. Attention should be paid to grammatical constraints. \\ \hline
        
        Verb-preposition 
        & A verb followed by a grammatically-constrained preposition, which attaches an indirect object to the verb.
        & \specialcell{{\em morph into} \\{\em shield against}}
        & Replace full MWE with a simpler MWE of the same syntactic pattern. Ensure grammaticality of the resulting simplification. 
        \\ \hline
        
        Verb-noun(-preposition) 
        & Conventionalized MWE where the syntactic head is a verb with a dependent noun that may attach further preposition.
        & \specialcell{{\em provides access to}\\ {\em bid farewell}}
        & Replace full MWE with a simpler word or MWE, taking care of grammatical constraints.
        \\ \hline
        
        Support verb 
        & Lexicalized constructions with light verbs ({\em make}, {\em take}, etc.).
        &\specialcell{ {\em make clear}\\ {\em has taken steps}}
        & Replace full MWE with a simpler verb.
        \\  \hline
        
        PP modifier 
        & Conventionalized phrase with a preposition as its syntactic head. 
        & \specialcell{{\em upon arrival} \\{\em within our reach}}
        & Simplification may involve elaboration using a relative clause.
        \\ \hline
        
        Coordinated phrase 
        & Lexicalized phrases involving coordination.
        & \specialcell{{\em shock and horror}\\ {\em import and export}}
        & Simplification would typically involve replacement of the whole MWE; additional explanation may need to be provided in case of fixed phrases.
        \\ \hline
        
        Conjunction / Connective 
        & An MWE which is used to connect two parts of a sentence.
        & \specialcell{{\em thus far}\\ {\em according to}}
        & May require syntactic rather than lexical simplification.
        \\ \hline
        
        Semi-fixed VP 
        & Conventionalized verbal phrase which allows some degree of lexical variation (e.g. inflection, variation in reflexive form, and determiner selection).
        & \specialcell{{\em flexed $<$their$>$ muscles}\\ {\em close $<$the$>$ deal}}
        & The phrase and non-fixed unit may require simplifying separately. Care should be taken when simplifying the phrase to ensure agreement with the non-fixed unit.
        \\ \hline
        
        Fixed phrase 
        & A frequent, lexicalized, non-compositional phrasal expression; this category also includes borrowed expressions 
        & \specialcell{{\em conflict of interest}\\ {\em the tide has turned} \\ {\em et al.}}
        & As such MWEs are typically idiomatic, they may require an explanation to be given, rather than a simplification.
        \\ \hline
        
        Not MWE 
        & A special category for annotated ``phrases" that are not MWEs proper (sentence fragments, fully transparent expressions, etc.)
        & \specialcell{{\em vehicle rolled over}\\ {\em IP address is blocked}}
        & These should not be simplified as a single unit, but instead simplified using other appropriate strategies (e.g., on a word-by-word basis).
        \\ \hline
        
        Not MWE but contains MWE(s) 
        & A ``phrase" that is not an MWE proper as a whole, but contains an MWE as a sub-unit. 
        &\specialcell{ {\em collapsed \underline{property sector}} \\ {\em \underline{interior ministry} troops} }
        & The MWE sub-unit should be classified and simplified according to the categories above.
        \\ \hline
\end{tabular}
\caption{Classes of MWEs annotated in our data}
\label{tab:categories}
 \end{center}
\end{table*}

\subsection{Annotation Protocol}

Annotation was performed in two phases: first, $1000$ instances were annotated by all three annotators over a series of rounds. The rounds comprised of annotating $100$, $200$, $300$ and $400$ instances. After each round, an inter-annotator agreement (IAA) was evaluated using Fleiss' kappa ($\kappa$)~\cite{fleiss1981}. The annotators met to discuss and resolve disagreements: in the majority of cases, $2$ out of $3$ annotators agreed. Disagreements were resolved to produce a single gold standard annotation for the final version of the dataset, resulting in the post-resolution IAA of $1.0$. Annotation guidelines were updated as necessary.  


The second phase consisted of individual annotation of the remainder of the dataset, split into three separate $1244$ instance chunks, by each of the annotators. After the corpus had been annotated we performed a number of consistency checks to minimize annotation errors:
\vspace{-0.3em}
\begin{itemize}
    \item We noted that it was often the case that the same phrase occurred in multiple contexts, with each case being annotated independently. To ensure annotation consistency, we checked whether such expressions had the same annotation throughout the dataset, and if any $2$ annotators disagreed on the label of an expression, the third annotator made a final decision. 
    \item We also noticed that some contexts were included in the dataset multiple times, producing a number of exact duplicates for the annotated phrases. To maximize consistency with the original data, we keep such exact duplicates in our dataset, making sure each of these expressions receives the same MWE annotation in all duplicate contexts.
    \item In addition, we checked all instances of {\tt Not MWE} to see if they contained any sub-unit which had been annotated elsewhere as an MWE. If this condition was met, we updated the label of such expression to {\tt Not MWE but contains MWE(s)}.
\end{itemize}

Table \ref{tab:stats} shows statistics, presenting the number of instances annotated in each round and pre-resolution IAA where applicable. We note that during the first $4$ rounds of joint annotation, we reach observed agreement of at least $0.70$ and $\kappa$ of $0.7145$ and higher, which amounts to substantial agreement~\cite{landis-koch-1977}, particularly given the high number of annotated categories in the data ($13$). Weighting the agreement for the number of instances in each round gives a final weighted agreement of $0.7978$ on the jointly annotated set. Individual fluctuations in agreement figures can be attributed to the growing number of examples from one round to the next one and heterogeneity of the randomized data splits.

\begin{table*}
\begin{center}
\begin{tabular}{|c|c|c|c c|}
      \hline
        \multicolumn{1}{|c|}{\multirow{2}{*}{\textbf{Phase}}} &
        \multicolumn{1}{c|}{\multirow{2}{*}{\textbf{Round}}} & \multicolumn{1}{c|}{\textbf{Number}} & \multicolumn{2}{c|}{\textbf{Agreement}} \\
        & & \textbf{of instances} & observed & $\kappa$ \\
      \hline
      & $1$ & $100$ & $0.7000$ & $0.7509$ \\
      First & $2$ & $200$ & $0.8342$ & $0.7779$ \\
      (joint annotation) & $3$ & $300$ & $0.7994$ & $0.7276$ \\
      & $4$ & $400$ & $0.8029$ & $0.7145$ \\ \hline
      Second & \multirow{2}{*}{$5$} & $1244$ & \multirow{2}{*}{-} & \multirow{2}{*}{-} \\
      (individual annotation) & & each & & \\ \hline
\end{tabular}
\caption{Statistics on the annotated dataset totalling $4732$ phrases}
\label{tab:stats}
 \end{center}
\end{table*}

\subsection{MWE Type Analysis}


Next, we analyze the distribution of various MWE types in data and draw conclusions about the most problematic MWE types for native and non-native readers. We stress that in the original data ``phrases" were identified by asking annotators to highlight sequences of words difficult to understand in context. Sequences of words with complexity score of $0$ in the binary setting and $0.0$ in the probabilistic setting represent simple ``phrases" not selected by any annotators as complex which were extracted to provide examples of the simple class. Since such ``phrases" were not explicitly annotated for complexity, and the procedure for their extraction from the data is not clearly defined, we do not include these cases in our analysis.\footnote{We have, nevertheless, provided MWE annotation for such cases and include them in the released dataset for consistency with the original data. We believe that including them in the statistical analysis of MWE type complexity will not be informative. Future research may look into more focused extraction of simple MWEs from this data.}

The frequencies of each annotation type in the full dataset combining both native and non-native reader annotations are shown in Table~\ref{tab:types}. The majority of the phrases that had been selected are {\tt not MWE} amounting to $46.09\%$ in the original data, and rising to $55.30\%$ when {\tt not MWE but contains MWE(s)} cases are taken into account. This shows that a vast majority of the sequences of words selected by the annotators in the original data are not MWEs. Instead, they are sequences of individually complex words that should be simplified independently.

The next most frequent types are {\tt MW compounds} and {\tt MW named entities} with $26.88\%$ and $10.50\%$ examples, respectively. At the same time, {\tt support verbs} and {\tt coordinated phrases} are the two least frequent categories with $7$ and $11$ examples in the whole dataset respectively. This corresponds to the observations of~\newcite{schneider-et-al-2014}.


After removing the randomized simple MWEs, we observe that the relative frequencies between annotation types do not change drastically, with only {\tt semi-fixed VP} and {\tt verb-preposition}, and {\tt verb-noun(-preposition)} and {\tt coordinated phrase} categories changing order in terms of frequency.

\begin{figure*}[h]
\begin{center}

\includegraphics[width=\textwidth]{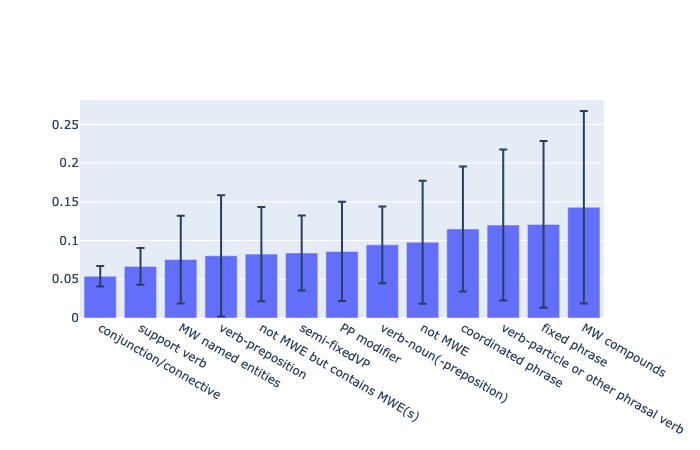}
\caption{The mean complexity (bar height) and standard deviation (error bar) of each MWE type}
\label{fig:bar}
\end{center}
\end{figure*}

We also investigate the correspondence between the MWE types and the complexity scores assigned to the instances of each type by the annotators, where the complexity scores represent the proportion of $20$ annotators who indicated that the expression is complex. Table \ref{tab:types} includes the mean complexity values for each MWE type, along with the standard deviation values, while Figure~\ref{fig:bar} visualizes these findings, with the MWE types ordered by their complexity. Overall, {\tt MW compounds} are the most complex type of MWEs, followed by {\tt fixed phrase} and {\tt verb-particle or other phrasal verb} categories. This trend corresponds to the degree of compositionality in the phrases: the rightmost extremity of the chart contains MWE types that are often semantically idiosyncratic. For instance, {\em  financial cushion} (annotated as {\tt MW compound} in our dataset), {\em the tide has turned} ({\tt fixed phrase}) or {\em staying put} ({\tt verb-particle or other phrasal verb}) are all non-compositional. The leftmost extremity of the chart covers phrases that may, to a certain degree, be compositional and semantically transparent: for instance, in combinations with {\tt support verbs} nouns are typically used in their usual sense, while verb meanings appear to be bleached, rather than idiomatic~\cite{sag-et-al-2002}, which might help readers understand these types of phrases. We note that the complexity of the {\tt MW named entities} type is a matter of world knowledge and varies widely between individuals, explaining the relatively low overall complexity for this type with high standard deviations.

\begin{figure*}
\centering
\includegraphics[width=\textwidth]{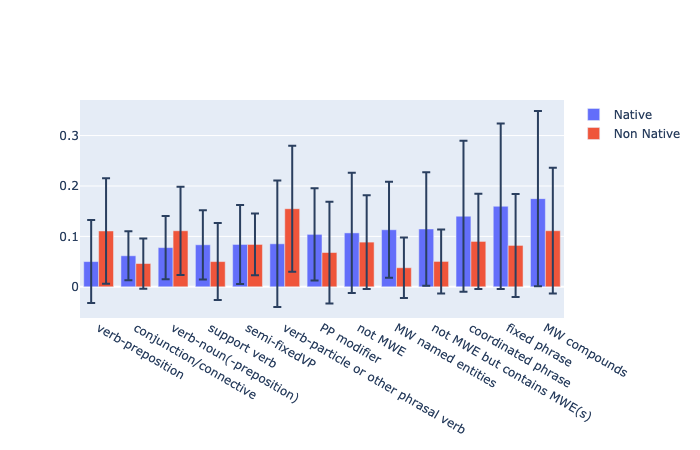}
\caption{A comparison of the Native Annotator's complexity labels vs. Non-native Annotator's complexity labels for each MWE category}
\label{fig:nativeVNonNative}
\end{figure*}

Figure \ref{fig:nativeVNonNative} complements these findings by highlighting the differences in complexity annotation between native and non-native readers. We note that non-native readers find verbal expressions in {\tt verb-preposition} and {\tt verb-particle or other phrasal verb} constructions noticeably more challenging.

These results demonstrate that there is considerable variation in complexity between MWE types, and this further motivates our research into incorporation of MWE types into a lexical complexity assessment system.

\begin{table*}[t]
\begin{center}
\begin{tabular}{|l|r r|r r|r r|}
      \hline
      \multicolumn{1}{|c|}{\multirow{2}{*}{\textbf{MWE Type}}} & \multicolumn{2}{c|}{\textbf{Original}} & \multicolumn{4}{c|}{\textbf{Complex only}} \\
       & Total & \% & Total & \multicolumn{1}{c}{\%} & Mean & Std\\
      \hline
        not MWE  &                              2181 & 46.09 & 1665 & 44.45 & 0.101 & 0.098\\
        MW compounds    &                       1272 & 26.88 & 1131 & 30.19 & 0.145 & 0.143\\
        MW named entities &                     497  & 10.50 & 365  & 9.74  & 0.077 & 0.075\\
        not MWE but contains MWE(s) &           436  & 9.21  & 300  & 8.01  & 0.088 & 0.083\\
        verb-particle or other phrasal verb &   119  & 2.51  & 102  & 2.72  & 0.127 & 0.120\\
        fixed phrase &                          72   & 1.52  & 67   & 1.79  & 0.119 & 0.121\\
        semi-fixedVP &                          39   & 0.82  & 25   & 0.67  & 0.083 & 0.084\\
        verb-preposition &                      34   & 0.72  & 28   & 0.75  & 0.078 & 0.080\\
        PP modifier &                           33   & 0.70  & 25   & 0.67  & 0.087 & 0.086\\
        conjunction/connective &                16   & 0.34  & 13   & 0.35  & 0.054 & 0.054\\
        verb-noun(-preposition) &               15   & 0.32  & 9    & 0.24  & 0.115 & 0.094\\
        coordinated phrase &                    11   & 0.23  & 10   & 0.27  & 0.125 & 0.115\\
        support verb &                          7    & 0.15  & 6    & 0.16  & 0.070 & 0.067\\
      \hline
\end{tabular}
\caption{The frequency and complexity of each MWE type, full dataset}
\label{tab:types}
 \end{center}
\end{table*}

\section{MWE Complexity Assessment Systems} \label{sec:baseline}
Evaluating the complexity of MWEs is a two step process, as the initial identification that an expression is an MWE is required prior to predicting its complexity. We leave the MWE identification step to future research. Instead, we operate on the assumption that an oracle system has identified the MWEs in our data, and build a lexical complexity system whose goal is to assign a complexity score to the identified MWEs. The complexity assessment system is trained and evaluated on the $2551$ phrases that are annotated as an MWE in our dataset. In the binary setting, only $470$ have label $0$ and the rest are annotated as complex with label $1$ so we run more fine-grained experiments under the probabilistic setting, which represents the complexity of a phrase on a scale of $[0.0...0.70]$,\footnote{The uppper bound on this scale reflects the fact that at most $14$ annotators agreed that a particular phrase is complex.} representing the proportion of $20$ annotators that found a phrase complex. The MWE complexity assessment system is a supervised feature-based model.

\subsection{Features}
Our baseline complexity assessment system relies on $6$ features. We include two traditional features found to correlate highly with word complexity: {\em length} and {\em frequency}. These are adapted for phrases by considering (1) the number of words instead of the number of characters for {\em length}, and (2) using the average frequency of bigrams within the phrase, which is calculated using the Corpus of Contemporary American English \cite{davies2009385+} for {\em frequency}. The second category of features focuses on the complexity of words contained within the MWE. We use an open source system of \newcite{gooding-kochmar-2019-complex} to tag words with a complexity score, whereupon the highest word complexity within the phrase as well as the average word complexity are included as features. The source genre of phrases is included in the feature set, as genre acts as a proxy of world knowledge. Finally, the feature of primary importance in experimentation is that of MWE type, derived from our MWE-annotated dataset. Table \ref{tab:features} illustrates the feature set for the phrase \textit{sledge hammers}. 


\begin{table}[h]
\centering
\begin{tabular}{l|l}
        & \textit{sledge hammers} \\ \hline
MWE     & MW Compounds   \\
Length  & 2              \\
Freq    & 39             \\
Max CW  & 0.70           \\
Mean CW & 0.60           \\
Genre   & News          
\end{tabular}
\caption{Feature set for \textit{sledge hammers}}
\label{tab:features}
\end{table}

\subsection{System Implementation}
We model the task of complexity prediction as a regression task. Therefore, we apply a set of standard regression algorithms from the {\tt scikit-learn}\footnote{\url{https://scikit-learn.org}} library. Model predictions are rounded to the closest $0.05$ interval. The best performing model found during preliminary experimentation uses a Multi-layer Perceptron regressor with $6$ hidden layers and the {\tt lbfgs} optimiser, used due to the size of the dataset. 

\subsection{Results}
\begin{table}[]
\centering
\begin{tabular}{lll}
\multicolumn{1}{l|}{Test Set}   & \multicolumn{2}{c}{MAE}      \\ \hline
\multicolumn{1}{l|}{} & \multicolumn{1}{l|}{{\sc Camb}} & {\sc Our System}   \\ \hline
\multicolumn{1}{l|}{{\sc News} (131)}       & \multicolumn{1}{l|}{0.0748}  &  \underline{0.0603}\\
\multicolumn{1}{l|}{{\sc Wikipedia} (78)}   & \multicolumn{1}{l|}{0.0744}   & \underline{0.0691} \\
\multicolumn{1}{l|}{{\sc WikiNews (80)}}  & \multicolumn{1}{l|}{\underline{0.0325}}   & 0.0369 \\ 

\end{tabular}
\caption{MAE scores achieved by our system and baseline systems}
\label{tab:results}
\end{table}

We compare our results to two baselines: first, we compare our results to the strategy used by the winning shared task system {\sc CAMB}~\cite{gooding-kochmar-2018-camb} where all phrases are simply assigned the complexity value of $0.05$. The second baseline is based on outputting the most common probabilistic label observed in the training data: this typically always results in a complexity value of $0.05$, however for some test sets such as {\sc WikiNews} this would be $0.00$. These baselines are highly competitive as $1074$ of the $2551$ examples have a probabilistic score of $0.05$, with $61\%$ of MWEs having a value of $0.00$ or $0.05$. We use Mean Absolute Error (MAE) as our evaluation metric, following the $2018$ Shared Task~\cite{yimam-etal-2018-report}. 

We report the results on the MWE portion of the $2018$ shared task test sets at the top of Table \ref{tab:results} alongside the baseline {\sc camb} system. Our system achieves lower absolute error on both {\sc News} and  {\sc Wikipedia} test sets, but not on {\sc WikiNews} test set (the best results are underlined in Table \ref{tab:results}). It is worth noting that the distribution of probabilistic scores in this test set is highly skewed, with $79$\% having scores of $0.05$ or $0.00$ and the highest score in the dataset being only $0.35$.

\begin{table}[]
\centering
\begin{tabular}{l|l}
MAE & Feature group     \\ \hline
0.0577 & All \\ \hline
\textbf{0.0673}  & - \textit{MWE} \\
0.0617  & - \textit{Genre} \\
0.0602   & - \textit{Mean CW} \\
0.0580  & - \textit{Max CW} \\
0.0584   & - \textit{Length} \\
0.0581  & - \textit{Frequency} \\ \hline
0.0641 & {\sc Baseline}
\end{tabular}
\caption{5-fold cross-validation experiments and ablation tests on the entire dataset using our system}
\label{tab:ablations}
\end{table}

Table \ref{tab:ablations} includes evaluation on the entire dataset using $5$-fold cross-validation. To investigate the informativeness of features we perform ablation tests by excluding each feature and observing the impact on performance. Features are listed in order of their impact. The most informative feature is the type of MWE (highlighted in bold), followed by the genre. These features contribute to the largest increase in MAE. The comparative baseline presented at the bottom of the table uses the mode label from the training set. 

The same set experiments are also performed on native and non-native probabilistic annotations, with results presented in Table \ref{tab:native_non_results}. The annotations of each group are considered separately during training and testing. We note that there is a considerable difference in annotations: for instance, native annotations cover the full scale of $[0.0...1.0]$, while non-native annotations fall between $[0.0...0.8]$, with both sets in this case having a step of $0.1$ which represents one annotator. The most informative features for each group differ, with the best results for native annotators being obtained without frequency and length information. For the native group the most informative features are the type of MWE, word complexity features and genre. However, for the non-native group the best results are achieved when using all available features. Intuitively, this makes sense as non-native readers rely on brevity and frequency when learning vocabulary. The system trained to predict non-native complexity outperforms the native system, and both systems are able to beat respective baselines. 

\begin{table}[]
\centering
\begin{tabular}{l|ll}
           & MAE      & \textit{-MWE}   \\ \hline
Native     & 0.0936   & 0.0971          \\
{\sc Baseline}   & \multicolumn{2}{c}{0.1185} \\ \hline
Non-native & 0.0698   & 0.0737          \\
{\sc Baseline}   & \multicolumn{2}{c}{0.0823}
\end{tabular}
\caption{MAE scores obtained by our system on native and non-native complexity annotations}
\label{tab:native_non_results}
\end{table}


\section{Discussion}

Our results show that the inclusion of MWE type labels improves complexity estimation. Using an ablation study, we find that the category of MWE is the most informative feature when predicting probabilistic complexity (see Table \ref{tab:ablations}). We observe in the dataset that the mean complexity varies across categories. For instance {\tt MW compounds} has a mean probabilistic value of $0.127$ compared to $0.044$ for the {\tt conjunctive/connective} category. The variation in mean complexity values across categories indicates the average difference in difficulty for the readers. 

The performance also differs between systems trained to predict native vs non-native probabilistic complexity scores. Whilst MWE type is informative in both cases, the best performing feature sets are considerably different. Notably, frequency and length are helpful when predicting complexity scores for non-native readers but not when considering the native case. The overall results on native complexity prediction are worse than those for the non-native group, despite the inter-annotator agreement in the original data being higher for the native reader group~\cite{yimam-etal-2017-cwig3g2}. Further work to identify which features and systems work best for each group is needed. Regarding the MWE type, the dataset illustrates differences across the mean complexity depending on the group of annotators. For instance, the {\tt MW compounds} category has an average probabilistic complexity of $0.156$ for native readers and $0.098$ for non-native ones. This is the highest mean for both groups across all categories suggesting that MW compounds can be universally challenging. However, in addition to the findings presented in Figure \ref{fig:nativeVNonNative}, there are clear group differences even in the types of MW compounds that readers find complex. Table \ref{tab:most_complex_MWC} illustrates two such examples:

\begin{table}[h]
\centering
\begin{tabular}{l|ll}
\textit{}             & \textit{Native} & \textit{Non-Native} \\ \hline
\textit{Pool report}  & 1.0             & 0.3                 \\
\textit{Edit Warring} & 0.3             & 0.8     
\end{tabular}
\caption{Complexity annotation differences on MW compounds}
\label{tab:most_complex_MWC}
\end{table}

\section{Conclusion}

We have shown that the probabilistic complexity of MWEs varies according to the type of MWE. In addition to this, the types of MWEs that native and non-native speakers find to be complex also vary widely. In our experiments, we have developed baseline regressors that attempt to predict the complexity of MWEs based on a number of hand-crafted features. We show that MWE type is the most informative feature when trying to predict the complexity of MWEs. We have not addressed the wider task of identifying MWEs from free text, or their types, however our corpus could be used as a starting point to do so.

\section*{Acknowledgements}

The first author's research is supported by Cambridge Assessment, University of Cambridge, via the ALTA Institute. We are grateful to the anonymous reviewers for their valuable feedback.

\section{Bibliographical References}

\bibliographystyle{lrec}
\bibliography{lrec2020W-xample}


\end{document}